\documentclass[10pt]{IEEEtran}
\usepackage[noadjust]{cite}
\usepackage{amsmath}
\usepackage{amssymb}
\usepackage{amsthm}
\usepackage{mathtools}
\usepackage{bbm}
\usepackage{algorithm}
\usepackage[noend]{algpseudocode}
\usepackage{bm}
\usepackage{comment}
\usepackage{tabularx,booktabs}
\usepackage[capitalize]{cleveref}
\usepackage{eqnarray}
\usepackage{float}
\usepackage{pgfplots}
\pgfplotsset{compat=1.15}
\usepgfplotslibrary{fillbetween}
\usetikzlibrary{patterns,arrows,plotmarks}
\usepgfplotslibrary{groupplots}
\pgfdeclarelayer{background}
\pgfsetlayers{background,main}
\usetikzlibrary{automata,positioning}
\usetikzlibrary{decorations}
\usetikzlibrary{shapes.arrows}
\usetikzlibrary{tikzmark}
\usetikzlibrary{calc}
\usetikzlibrary{decorations.markings}
\algrenewcommand\algorithmicindent{10pt}
\usepgfplotslibrary{colorbrewer}
\usepgfplotslibrary{groupplots}
\usepackage[caption=false]{subfig}
\usepackage{multirow,multicol}

\usepackage{pifont}
\newcommand{\cmark}{\ding{51}}%
\newcommand{\xmark}{\ding{55}}%
\definecolor{limegreen}{RGB}{50,205,50}

\newtheorem{definition}{Definition}

\newcommand{\mc}[1]{\mathcal{#1}}   
\newcommand{\mb}[1]{\mathbf{#1}}    
\DeclareMathOperator*{\argmax}{arg\,max}    

\usepackage[acronym]{glossaries}

\newacronym{api}{API}{Alternate Policy Iteration}

\newacronym{sfr}{SFR}{Successor Feature Representation}
\newacronym[\glslongpluralkey={Constrained Markov Decision Processes}]{cmdp}{CMDP}{Constrained Markov Decision Process}
\newacronym{cnn}{CNN}{Convolutional Neural Network}
\newacronym{cr}{CR}{Compression Ratio}
\newacronym{dccp}{DCCP}{Disciplined Convex-Concave Programming}
\newacronym{decpomdp}{Dec-POMDP}{Decentralized POMDP}
\newacronym{dnn}{DNN}{Deep Neural Network}
\newacronym{drl}{DRL}{Deep RL}
\newacronym{dqn}{DQN}{Deep Q-Networks}

\newacronym{fnn}{FNN}{Feed-Forward Neural Network}

\newacronym{ge}{GE}{Gilbert-Elliott}
\newacronym{gpmdp}{GP-MDP}{Group Policy MDP}
\newacronym{goc}{GoC}{Goal-oriented Commmunication}

\newacronym{hr3l}{HR3L}{Homomorphic Robust Remote Reinforcement Learning}

\newacronym{ibr}{IBR}{Iterated Best Response}

\newacronym[\glslongpluralkey={Markov Decision Processes}]{mdp}{MDP}{Markov Decision Process}
\newacronym[\glslongpluralkey={Remote Markov Decision Processes}]{rmdp}{RMDP}{Remote Markov Decision Process}
\newacronym{mbrl}{MBRL}{Model-Based Reinforcement Learning}
\newacronym{mpc}{MPC}{Model Predictive Control}

\newacronym{ne}{NE}{Nash Equilibrium}
\newacronym{pec}{PEC}{Packet Erasure Channel}
\newacronym[\glslongpluralkey={Policy-Level Constrained Markov Decision Processes}]{plcmdp}{PLC-MDP}{Policy-Level Constrained Markov Decision Process}
\newacronym[\glslongpluralkey={Partially Observable Markov Decision Processes}]{pomdp}{POMDP}{Partially Observable Markov Decision Process}
\newacronym{porl}{PORL}{Partially Observable Reinforcement Learning}
\newacronym{poa}{PoA}{Price of Anarchy}
\newacronym{ppo}{PPO}{Proximal Policy Optimization}

\newacronym{qp}{QP}{Quadratic Program}
\newacronym{qclp}{QCLP}{Quadratically Constrained Linear Program}

\newacronym{rl}{RL}{Reinforcement Learning}
\newacronym{rb}{RB}{Rollout Buffer}
\newacronym{rrl}{RRL}{Remote Reinforcement Learning}
\newacronym{rtt}{RTT}{Round Trip Time}

\newacronym{ssl}{SSL}{Self-Supervised Learning}
\newacronym{sde}{SDE}{State-Dependent Exploration}

\newacronym{jscc}{JSCC}{Joint Source Channel Coding}

\definecolor{msecol}{HTML}{0011af}
\definecolor{avgcol}{HTML}{8819a0}
\definecolor{varcol}{HTML}{bf418d}
\definecolor{maxcol}{HTML}{e37076}
\definecolor{cntcol}{HTML}{f9a256}
\definecolor{mafcol}{HTML}{ffd700}

\definecolor{lightgray}{HTML}{999999}
\definecolor{color0}{HTML}{00429D}
\definecolor{color1}{HTML}{844D99}
\definecolor{color2}{HTML}{C3608E}
\definecolor{color3}{HTML}{EF8078}
\definecolor{color4}{HTML}{FFB047}

\definecolor{col1}{HTML}{ffb047}
\definecolor{col2}{HTML}{e69267}
\definecolor{col3}{HTML}{c7777c}
\definecolor{col4}{HTML}{a0618b}
\definecolor{col5}{HTML}{6e4f96}
\definecolor{col6}{HTML}{00429d}

\definecolor{c14}{HTML}{ffb047}
\definecolor{c24}{HTML}{d27f76}
\definecolor{c34}{HTML}{915a8f}
\definecolor{c44}{HTML}{00429d}

\begin{document}

\title{Robust Remote Reinforcement Learning over Unreliable Communication Channels using Homomorphic State Encoding}

\author{Pietro Talli,~\IEEEmembership{Graduate Student Member,~IEEE,} Federico Mason,~\IEEEmembership{Member,~IEEE,}\\Federico Chiariotti,~\IEEEmembership{Senior Member,~IEEE,}  and Andrea Zanella,~\IEEEmembership{Senior Member,~IEEE}%
\thanks{All authors are with the Department of Information Engineering, University of Padova, via Gradenigo 6B, 35131 Padua, Italy (emails: \{tallipietr, masonfed, chiariot, zanella\}@dei.unipd.it). This work was partly supported by the SNS JU project MAGIC-6G under Horizon Europe grant no. 101292933.}
}
\maketitle
\thispagestyle{empty} 
\pagestyle{empty}     

\begin{abstract}
Traditional \gls{rl} frameworks generally assume that the agent perceives the state of the underlying Markov process instantaneously and then takes actions accordingly. 
If the agent cannot directly observe the process, but rather receives state updates from a remote sensor over a lossy and/or delayed channel, it may be forced to operate with partial and intermittent information.
In recent years, numerous learning architectures have been proposed to manage \gls{rl} with imperfect or remote feedback; however, they offer solutions tailored to specific use cases, often with a substantial computational and communication burden.
To address these limitations, we propose a novel learning architecture, named \textit{\gls{hr3l}}, that enables the distributed training of \gls{rl} agents over unreliable communication channels without the need to exchange gradient information.
Our experimental results demonstrate that \gls{hr3l} significantly outperforms the state-of-the-art methods in terms of sample efficiency, leading to faster training and reduced communication overhead.
In addition, we show that \gls{hr3l} can adapt to different scenarios, including packet loss, delayed transmissions, and bandwidth limitations, without experiencing significant performance degradation.
\end{abstract}
 
\glsresetall

\section{Introduction}

\IEEEPARstart{I}{n} recent years, 
the \gls{rl} paradigm has captured a broad interest from both the scientific and industrial communities, as it enables the discovery of efficient decision-making policies to handle a wide variety of scenarios~\cite{sutton1998reinforcement}. 
Under the assumption that the target environment evolves as a Markov process, \gls{rl} enables the refinement of any planning or control task that can be decomposed into sequences of decisions.
The fundamental requirement for the learning process is that the agent receives feedback to assess the effects of its actions and predict long-term performance under the current policy.
Crucially, this feedback loop relies on a continuous flow of information between the environment and the agent, making performance highly sensitive to the reliability of the underlying communication process.

The above aspect becomes critical in the so-called \textit{\gls{rrl}} scenarios, such as edge/cloud computing or remote control applications, where the \gls{rl} agent may not directly observe the environment, relying instead on the information collected and transmitted back by remote nodes.
In real-world communication networks, factors such as packet loss, transmission delay, limited bandwidth, and node failure can severely compromise the feedback loop, thus affecting the learning of the optimal policy \cite{roig2020remote}.
This may constitute a significant barrier to the realization of \gls{rl} systems in remote control applications~\cite{yahmed2023deploying}.
Importantly, \gls{rrl} naturally integrates with the concept of \gls{goc}~\cite{strinati20216g}, in which transmission decisions are strictly functional to achieve system-wide objectives.

\subsection{Related Work}
In the literature, the optimization of \gls{rrl} systems with unreliable communication has been studied through two main approaches: (i), modeling the environment and adapting the control policy to cope with missing information, or (ii), jointly adapting communication and control policies while accounting for channel unreliability.
This taxonomy guides the following review of the literature, as summarized in Table~\ref{tab:soa}. 

\begin{table*}[htb]
\caption{Summary of existing approaches in the literature.}
\centering
\begin{tabular}{p{0.05\linewidth}|p{0.17\linewidth}|>{\centering\arraybackslash}p{0.07\linewidth}|>{\centering\arraybackslash}p{0.07\linewidth}|>{\centering\arraybackslash}p{0.07\linewidth}|>{\centering\arraybackslash}p{0.07\linewidth}|>{\centering\arraybackslash}p{0.07\linewidth}|>{\centering\arraybackslash}p{0.07\linewidth}|>{\centering\arraybackslash}p{0.07\linewidth}}
\toprule
\multicolumn{2}{c|}{\textbf{Approach}} & \textbf{Model learning}   & \textbf{Long-term prediction} & \textbf{Low computational cost} & \textbf{Low communication overhead} & \textbf{Low sample complexity} & \textbf{Unreliable channels} & \textbf{Time-varying channels} \\ \midrule
\multirow{3}{*}{\textbf{Control}} & MPC~\cite{campo1987robust} & \textcolor{red}{\xmark} & \textcolor{red}{\xmark} & \textcolor{limegreen}{\cmark} & \textcolor{limegreen}{\cmark} & \textcolor{limegreen}{\cmark}& \textcolor{limegreen}{\cmark} & \textcolor{red}{\xmark}\\
& MBRL~\cite{moerland2023model} & \textcolor{red}{\xmark} & \textcolor{red}{\xmark} & \textcolor{red}{\xmark} & \textcolor{limegreen}{\cmark} & \textcolor{limegreen}{\cmark} & \textcolor{limegreen}{\cmark} & \textcolor{red}{\xmark}\\
&PORL~\cite{kaelbling1998planning,chen2024efficient,smallwood1973optimal,karamzade2024reinforcement, howson2023delayed}& \textcolor{limegreen}{\cmark} &  \textcolor{limegreen}{\cmark} & \textcolor{red}{\xmark} & \textcolor{limegreen}{\cmark} & \textcolor{red}{\xmark} & \textcolor{limegreen}{\cmark} & \textcolor{red}{\xmark}\\
\midrule
\multirow{4}{*}{\textbf{Joint}} & Pull-based RRL~\cite{83e8fe62,wang2024ocmdp,bellinger2023dynamic} & \textcolor{limegreen}{\cmark} &\textcolor{limegreen}{\cmark} &\textcolor{limegreen}{\cmark} & \textcolor{red}{\xmark} & \textcolor{green}{\xmark}&\textcolor{limegreen}{\cmark} & \textcolor{red}{\xmark}\\
& Push-based RRL~\cite{tung2021effective,mason2024multi,talli2025pragmatic} & \textcolor{limegreen}{\cmark} &\textcolor{limegreen}{\cmark} &\textcolor{limegreen}{\cmark} & \textcolor{red}{\xmark} & \textcolor{red}{\xmark}&\textcolor{limegreen}{\cmark} & \textcolor{red}{\xmark}\\
& JSCC~\cite{lyu2024semantic, shao2021learning}& \textcolor{limegreen}{\cmark} &\textcolor{red}{\xmark} &\textcolor{red}{\xmark} & \textcolor{red}{\xmark} & \textcolor{red}{\xmark}&\textcolor{limegreen}{\cmark} & \textcolor{red}{\xmark}\\
&HR3L (ours) & \textcolor{limegreen}{\cmark} & \textcolor{limegreen}{\cmark} & \textcolor{limegreen}{\cmark} & \textcolor{limegreen}{\cmark} & \textcolor{limegreen}{\cmark} & \textcolor{limegreen}{\cmark} & \textcolor{limegreen}{\cmark} \\  
                    \bottomrule
\end{tabular}\label{tab:soa}
\end{table*}

\subsubsection{Control-Policy Adaptation}
A possible approach to handle an unreliable feedback channel in remote control scenarios is to leverage \gls{mpc}, where the control agent uses a mathematical model of the environment to predict states whose observation is missed or delayed~\cite{campo1987robust}.  
However, accurately modeling complex, dynamic, and often time-varying environments in advance is often infeasible in real-world scenarios, which limits the reliability and applicability of \gls{mpc} solutions. 

A generalization of \gls{mpc} is provided by \gls{mbrl}, which allows the control agent to learn the environment model online, making the solution more practical in real-world scenarios.
Also in this case, the agent can simulate the trajectories of future states and actions to effectively handle missing or delayed observations~\cite{moerland2023model}.
However, \gls{mbrl} suffers from compounding errors, where inaccuracies in one-step predictions accumulate over time, leading to significant degradation in long-term performance.
Furthermore, learning accurate environment models, especially by prediction of the next state, is computationally intensive and ill-suited to edge network optimization, where devices have limited computational resources~\cite{janner2019trust}.  

Another direction is to represent the environment as a partially observable Markov decision process, where the controller models state observations as an independent, stochastic, and stationary process, thus casting the problem into a \gls{porl} framework~\cite{kaelbling1998planning}. 
In~\cite{chen2024efficient}, the \gls{porl} model has been used to prove that theoretical performance guarantees can be provided even when observations can be lost or delayed.  
However, the proposed solution requires exploring the entire \emph{belief space}~\cite{smallwood1973optimal}, that is, the set of all possible belief distributions over the problem's Markov state space, which is computationally infeasible for large-scale environments. 
Similar approaches~\cite{karamzade2024reinforcement, howson2023delayed} for delayed channels still require perfect knowledge of the environment model, with the same key limitation as \gls{mpc} and \gls{mbrl}.

\subsubsection{Joint Communication and Control Adaptation}
Control-based techniques address unreliable observations only on the control side, without considering compensation strategies adopted on the transmission side.
More advanced approaches jointly consider communication and control processes to mitigate the uncertainty of feedback, typically at the cost of a reward penalty~\cite{tung2021effective}, integrating the observations from control with the rich literature on \gls{goc}~\cite{strinati20216g}.
These approaches can be developed according to two main design principles: \textit{pull-based} methods, where a single agent is in charge of deciding both the communication and control policies; and \textit{push-based} methods, where the observation and transmission policies are decided by one agent, named \emph{transmitter}, while another agent, named \emph{receiver}, optimizes the control policy~\cite{pandey2025medium}. 

An example of pull-based \gls{rrl} model is given in~\cite{83e8fe62}, where the agent incurs a fixed penalty whenever it requests to observe the state of the environment. 
This model is extended in~\cite{wang2024ocmdp}, where the transmission cost depends on the agent’s specific decisions, which may result in different state observations.
The work in~\cite{bellinger2023dynamic} proposes a practical algorithm to address pull-based \gls{rrl} scenarios, allowing the agent to selectively observe parts of the state and determine how many time slots to remain silent before requesting new observations.
However, all pull-based frameworks share a fundamental limitation: the single decision-making agent does not have direct access to the environment's state, so that it has to make transmission and control decisions based on incomplete knowledge, often leading to suboptimal actions.

In contrast, in push-based \gls{rrl} scenarios, the communication decision is delegated to the observing agent (the transmitter), which has full knowledge of the current state of the system and can keep track of the beliefs of the control agent (the receiver).
Notable examples include~\cite{roig2020remote} and its extension in~\cite{tung2021effective}, where a two-agent \gls{rrl} architecture is used to explore a simple grid-world environment and performance is evaluated under ideal and noisy channel conditions. 
Recently, we developed a \gls{rrl} training framework based on an \emph{iterative best response} approach~\cite{mason2024multi}, where the control strategy is adapted to a fixed encoding scheme that, in turn, is optimized to better support control actions, repeating the process until the two policies converge.

Importantly, push-based \gls{rrl} scenarios can be optimized using \gls{jscc}, where the encoder and decoder assume the role of transmitter and receiver, respectively.
In recent years, \gls{jscc} led to promising encoding strategies for communication systems~\cite{lyu2024semantic, shao2021learning} that, however, share a critical issue: they require synchronous exchange of learning data between the controller and the observer through a differentiable channel.
Hence, whenever the control agent optimizes its policy, gradients must be propagated back to the observer agent, resulting in significant communication overhead during end-to-end training.
In particular, the channel must be differentiable to enable the transmitter to perform backpropagation through the channel, capturing the effect of control decisions on the encoding architecture.

In a recent work~\cite{talli2024push}, we have proved that push-based strategies are theoretically superior to pull-based strategies in \gls{rrl} scenarios.
However, most push-based solutions proposed in the literature assume differentiable channels without delays or packet losses, and are therefore ill-suited to deploy \gls{rrl} agents over unreliable communication channels.
Moreover, our subsequent work \cite{talli2025pragmatic} has demonstrated that push-based approaches can converge to local optima since they encounter coordination problems that result in complex game-theoretical interactions~\cite{xiao2012hierarchical}.
Identifying a globally optimal push-based \gls{rrl} strategy entails a complexity that scales exponentially with the state space, rendering the search computationally intractable for most practical applications.

Therefore, defining a push-based \gls{rrl} architecture that accounts for unreliable communication between the transmitter and receiver during end-to-end training remains an open problem, with no well-established solutions currently available in the literature.

\subsection{Novel Contributions}
In this work, we propose a novel framework for the training of  \gls{rrl} agents receiving observations through unreliable communication channels. 
Our system, named {\gls{hr3l}}, follows a push-based approach: the transmitter observes the state of the environment and decides whether to transmit it to the receiver, which then selects a control action and receives a reward. 
The core of \gls{hr3l} is the mechanism adopted by the transmitter to encode the environment information.
In particular, the transmitter exploits the theory of \gls{mdp} homomorphisms~\cite{ravindran2002model} to build a reduced \gls{mdp}, where the state and action spaces are designed to preserve only the features needed to maximize the expectation of the long-term reward. 
This allows the receiver to interact with a simplified \gls{mdp}, improving sample efficiency during end-to-end training while significantly reducing channel utilization.

Most importantly, unlike \gls{jscc} or similar approaches, \gls{hr3l} does not require propagating the gradients from the receiver to the transmitter side.
Instead, our framework organizes the training phase into a series of rounds, within which there is no data exchange between the observer and the control agents.
The only required signaling information is exchanged at the end of each round: in this phase, the receiver sends back the control decisions and instantaneous rewards to the transmitter, who then updates the receiver with the new version of the homomorphic \gls{mdp}. 
The two learning entities are thus trained in an almost fully distributed fashion, avoiding the need for synchronization and allowing the deployment of \gls{hr3l} in devices with limited resources.  
To our knowledge, this is the first \gls{rrl} training framework capable of scaling to arbitrary environment dimensions on the transmitter side while also supporting a distributed training phase.

Our experimental results demonstrate that the proposed \gls{hr3l} architecture significantly outperforms baseline \gls{rrl} strategies in handling packet loss and delayed feedback during transmitter–receiver data exchange.
When evaluated under stochastic delays derived from 5G communication traces, \gls{hr3l} experiences only one-third of the reward degradation observed in traditional \gls{rl} algorithms.
When implemented as a compression framework, \gls{hr3l} achieves a performance comparable to that of state-of-the-art learning-based compression strategies while reducing computational complexity by an order of magnitude, resulting in lower end-to-end delay.
Hence, in real-world communication networks characterized by capacity constraints and channel impairments, the performance gains provided by \gls{hr3l} become particularly significant.

Our contributions can be summarized as follows:
\begin{itemize}
    \item We define \gls{hr3l}, a novel learning architecture to train \gls{rrl} agents through an unreliable communication channel, without the need to transmit gradient data and ensure transmitter-receiver synchronization;
    \item We integrate \gls{hr3l} with a homomorphic representation of the environment that is used by the control agent to drive decisions over multiple steps, even when communication is sparse (due to compression) or absent (due to delay or packet loss);
    \item We evaluate \gls{hr3l} over a reference control test suite in multiple communication scenarios, characterized by impairments such as packet losses, delayed transmissions, and bandwidth constraints, showing that our method significantly improves performance and sample efficiency compared to state-of-the-art solutions.
\end{itemize}
An earlier version of this manuscript was presented as a conference paper~\cite{talli2026remote} and included an initial version of the \gls{hr3l} architecture considering low-dimensional inputs and no bandwidth constraints.
The complete architecture presented in this work can also deal with image-based inputs and bandwidth-constrained channels, as well as extending the evaluation by considering stochastic delays and variable impairments.

The remainder of this article is organized as follows:
Sec.~\ref{sec:model} presents the system model, formalizing the \gls{rrl} problem, and giving details of the communication between the learning agents.
Sec.~\ref{sec:homomorphic} provides a theoretical background on homomorphic \glspl{mdp}, which constitute the foundation of our framework.
Sec.~\ref{sec:method} then introduces the novel \gls{hr3l} architecture, while Sec.~\ref{sec:results} reports the settings of our experiments and analyzes the  performance of the proposed architecture against state-of-the-art solutions in a variety of tasks and communication conditions.
Finally, Sec.~\ref{sec:conclusion} provides conclusions and possible avenues for future work.

\section{System Model}
\label{sec:model}
We consider a \acrfull{rrl} push-based system with two agents, namely the \emph{transmitter} (or observer) and the \emph{receiver} (or controller).
The transmitter sends information regarding a certain process to the receiver, which controls the process to maximize a reward signal.
Importantly, the two agents communicate via a non-ideal channel, which may be characterized by capacity limitations, delays, and packet losses.
Consequently, both the communication policy of the transmitter and the control policy of the receiver must account for channel conditions.

\begin{figure}[t]
    \centering
    \include{schemes/system_model}
    \caption{Reference Remote Markov Decision Process (RMDP) scheme.}
    \label{fig:system_model}
\end{figure}

\subsection{Communication Model}
\label{subsec:cc}
We assume that time is discretized into slots $t=0,1,2,\ldots$ of a certain duration $\Delta t$.
As shown in Fig.~\ref{fig:system_model}, at each time slot, the transmitter observes the state $s_t$ of the \emph{environment} and encodes it into a \emph{message} $m_t$.
The message is then transmitted to the receiver, who performs a control action $a_t$ and obtains a reward signal $r_t$. 
Message transmission is modeled at the packet level as a limited-capacity delayed erasure channel, characterized by the following parameters:

\begin{itemize}
    \item The \textit{capacity} $C$ determines the maximum length $L$ of a message that can be transmitted in a time slot.
    \item The \textit{delay} $D_t$ is a random variable that determines the number $d_t=\left\lceil\frac{D_t}{\Delta t}\right\rceil$ of time slots after which the message $m_t$ sent at time $t$ reaches the receiver.
    In this work, we first consider $d_t$ to be deterministic, then address a scenario with stochastic delays. 
    \item The \textit{packet loss} is modeled via a Markov Gilbert-Elliot model, where messages are erased with probability $p_g$ or $p_b>p_g$ when the channel is in the \emph{good} or \emph{bad} states, respectively.
    The transition probabilities $P_{gb}$ from the good to the bad state and $P_{bg}$ from the bad to the good state determine the average probability of message erasure, given by $\bar p = \frac{p_b P_{gb} + p_g P_{bg}}{P_{bg}+P_{gb}}$, and the average duration of good and bad periods, equal to $1/P_{bg}$ and $1/P_{gb}$, respectively.
    This makes it possible to investigate the robustness of \gls{rrl} strategies to packet loss bursts.
    Notably, this channel model is a generalization of the Bernoulli erasure model with stationary loss probability that is typically considered in the existing literature on \gls{rrl} systems~\cite{chen2024efficient}. 
\end{itemize}

\subsection{The RRL Problem}
\label{subsec:rmdp}
In our system, the transmitter must learn to encode state observations into messages by selecting the features that are most useful to the control task of the receiver.
In turn, the receiver must tune the control actions according to the information received from the channel, which is determined by the communication policy.
The optimal communication and control policies can be learned by solving an infinite-horizon \gls{rmdp}, whose mathematical definition is provided below. 
\begin{definition}
\label{def:rrl}
A discounted infinite-horizon \gls{rmdp} is defined by a tuple $\langle \mc{S}, \mc{A}, \mc{M}, \mc{C}, P, r, \gamma \rangle$, where:
  \begin{itemize}
    \item $\mc{S}$ is the set of possible states;
    \item $\mc{A}$ is the set of possible actions;
    \item $\mc{M}$ is the set of possible messages;
    \item $\mc{C}$ is the communication channel model;
    \item $P:\mc{S}\times\mc{A}
    \to \Delta(\mc{S})$ is the state transition probability function;
    \item $r:\mc{S}\times\mc{A}\to\mathbb{R}$ is the reward function;
    \item $\gamma \in [0,1)$ is the exponential discount factor;
  \end{itemize}
\end{definition}
In the above definition, $\Delta(\Omega)$ denotes the set of all probability distributions with support in the discrete set $\Omega$.

As in traditional \gls{rl} scenarios, the state transition probability function $P$ describes the statistical evolution of the environment: given the state $s_t \in \mathcal{S}$ and the control action $a_t \in \mc{A}$ at slot $t$, the probability distribution of the next state is $P(s_t,a_t) \in \Delta(\mc{S})$.
With a slight abuse of notation, in the rest of the paper, we indicate the probability that the system moves to state $s_{t+1}\in \mc{S}$ at slot $t+1$ as $P(s_{t+1}|s_t, a_t)$.

In our system model, we assume that the receiver does not have perfect knowledge of the current state of the environment.
Instead, it  maintains a \emph{belief} $b_t\in \Delta(\mc{S})$ of the current state, based on the messages $\hat m_{1:t}$ received up to slot $t$ through the communication channel.\footnote{Note that $\hat{m}_{1:t}$ is a delayed version of the message process, so that $m_t$ might not be available yet.}
Therefore, the receiver's control policy $\pi$ does not operate directly in the state space, but in the \textit{belief} space $\Delta(\mc{S})$.

The optimal control policy $\pi^*$ must map the current belief $b_t$ into a probability distribution over the action space $\mathcal{A}$ to maximize the expected long-term discounted reward:
\begin{equation}
\label{eq:optimal_control}
    \pi^* = \argmax_{\pi:\Delta(\mc{S})\to\Delta(\mc{A})} \mathbb{E}_{a_t \sim \pi(b_t)} \left[ G(t) \right],
\end{equation}
where $G(t)=\sum_{t=0}^\infty \gamma^t r_t$ is the cumulative return, and the expectation is taken over the probability distribution defined in the action space given by the control policy $\pi$. 

On the sender side, the transmitter observes the state of the environment $s_t \in \mc{S}$ and encodes it in a message $m_t =\lambda(s_t) \in \mc{M}$ through an encoding function $\lambda:\mc{S}\to\mc{M}$.
Note that, in general, the encoding process may be lossy, so the state $s_t$ may not be fully recoverable from $m_t$.
The optimal encoding policy $\lambda^*$ must ensure that the receiver can maximize the expectation of $G(t)$, despite channel impairments.  
Therefore, the transmitter affects the reward indirectly through its effect on the receiver's beliefs about the system state. 

It is worth remarking that we assume that the reward signal $r_t$ is only available to the receiver at each time slot.
This restriction is realistic in remote control scenarios, though it is mostly ignored in the literature on \gls{rrl}, where the reward signal is assumed to be immediately available to all agents~\cite{tung2021effective}. 
In contrast, in our model, the receiver periodically shares the reward signals $r_{1:T}$ with the transmitter in an asynchronous fashion, i.e., at intervals lasting $T$ slots. 
These transmissions have loose time constraints, since a delayed delivery of the reward vector does not affect the learning process.
Therefore, we can  consider the feedback channel to be ideal, assuming that retransmission mechanisms can be deployed to recover any erased feedback message. 

\section{Homomorphic MDP Representation}
\label{sec:homomorphic}
Traditional value-based \gls{rl} methods set the goal of estimating the state-action value function $Q_{\pi}(s_t,a_t)$, representing the expected long-term reward when taking action $a_t \in \mc{A}$ in state $s_t \in \mc{S}$ and then following policy $\pi$:
\begin{equation}
\label{eq:q_fn}
    Q_{\pi}(s_t,a_t) = \mathbb{E}\left[r(s_t,a_t) + \sum_{\tau=t+1}^{\infty} \gamma^{\tau-t} r(s_{\tau},\pi(s_{\tau}))\right].
\end{equation}

In \gls{mbrl}, the agent explicitly computes the Q-value of each state-action pair  $(s, a) \in \mathcal{S} \times \mathcal{A}$ by building two functions $\hat{r}$ and $\hat{P}$ to estimate the reward signal and the transition probability, respectively~\cite{moerland2023model}.
These functions allow the agent to obtain a complete model of the environment and consequently estimate $Q_{\pi}(s_t,a_t)$ using the bootstrap method. 
On the other hand, a full environment representation is often difficult to learn and might contain features that are irrelevant to optimizing the agent policy.

A solution to the above issue is to map each state $s\in \mc{S}$ to a vector of significant features, using a basis function $\phi:\mc{S}\to\mathbb{R}^F$.
Ideally, $\phi(s)$ represents only the $F$ features of $s$ that are relevant to the learning problem.
We can follow the same approach for actions, defining a function $\alpha:\mc{A}\to\mathbb{R}^A$ that encodes each action $a \in \mathcal{A}$ to a vector of $A$ features. 
Therefore, the optimal policy can be learned without processing $s$ and $a$, but their encoded versions $\phi(s)$ and $\alpha(a)$, strongly reducing the training effort~\cite{hafner2023mastering}.

The foundation of this approach is given by the concept of \textit{homomorphic \gls{mdp}}~\cite{van2020mdp}, recalled below.
\begin{definition}
Two \glspl{mdp}, $\mathcal{P}=\langle \mc{S}, \mc{A}, P, r, \gamma \rangle$ and $\mathcal{P}'=\langle \mc{S}', \mc{A}', P', r', \gamma \rangle$, are homomorphic if and only if there exist two subjective maps $\phi:\mc{S}\to\mc{S}'$ and $\alpha:\mc{A}\to\mc{A}'$ such that
\begin{align}
    r'(\phi(s), \alpha(a)) & = r(s,a),\,\forall s\in\mc{S}, a\in\mc{A};\\
    P'(\phi(s^\prime)|\phi(s), \alpha(a)) & =P(s^\prime| s, a),\,\forall s, s^\prime\in\mc{S}, a\in\mc{A}.
\end{align}
\end{definition}
This concept can be extended to non-exact homomorphisms, where a theoretical error bound is set for the approximation of the state and action spaces~\cite{ravindran2004approximate}.
The problem is then to design functions $\phi$ and $\alpha$ to minimize the homomorphism distance.

An implicit implementation of the \gls{mdp} homomorphism theory is provided by \gls{sfr}, which was first conceived in~\cite{dayan1993improving} and was further developed in~\cite{borsa2018universal, chua2024learning}.
This technique enables the design of state and action representations aligned with long-term consequences associated with the agent's policy, not just immediate observations, overcoming the issue of compounding errors. 

In particular, \gls{sfr} allows us to decompose the reward as a linear combination of feature components:
\begin{equation}
r(s,a)=\begin{pmatrix}\phi(s)&\alpha(a)\end{pmatrix}^\intercal \mb{w},
\end{equation}
where $\begin{pmatrix}\phi(s)&\alpha(a)\end{pmatrix} \in \mathbb{R}^{(F+A)\times 1}$ is the column vector given by the concatenation of $\phi(s)$ and $\alpha(a)$, while $\mb{w} \in \mathbb{R}^{(F+A)\times 1}$ is a column vector that represents the contribution of each feature to the reward. 

We define the \gls{sfr} of $Q_{\pi}(s_t,a_t)$ as a column-vector of the same dimension as $\mb{w}$, computed as
\begin{equation}
\setlength\arraycolsep{2pt}
\psi_\pi(s_t,a_t) = \mathbb{E}_{a_t \sim \pi(s_t)} \left[ \sum_{\tau=t}^{\infty} \gamma^{\tau-t}     
    \begin{pmatrix}\phi(s_t)&\alpha(a_t)\end{pmatrix}    
    \!\mid\! s_t,a_t \right].
\end{equation}
Hence, the original Q-values can be retrieved as a linear combination of the elements of $\psi_{\pi}(s_t,a_t)$, weighted by $\mb{w}$:
\begin{equation}
\label{eq:sfr}
    Q_\pi(s_t,a_t) = \psi_\pi(s_t,a_t)^\intercal \mb{w}.
\end{equation}
As shown in~\cite{fujimoto2025towards}, this decomposition creates a versatile training framework to solve any \gls{rl} problems without the need for hyperparameter tuning.

\section{Learning Architecture}
\label{sec:method}

In this section, we present \acrfull{hr3l}, a distributed learning architecture to effectively train and deploy \gls{rrl} agents over unreliable communication channels.
Following the model from Sec.~\ref{sec:model}, the architecture includes two learning units: the transmitter, which extracts relevant features from the state of the system, and the receiver, which solves the control task by exploiting the information received through the channel.

Importantly, \gls{hr3l} optimizes the encoding policy $\lambda$ and the control policy $\pi$ of the receiver in a fully distributed manner, avoiding the need to perform backpropagation through a differentiable channel or to perform synchronized gradient updates.
Specifically, the transmitter and receiver exchange \textit{learning data} only at the end of each training round; no feedback is transmitted within the round itself.

\subsection{Transmitter Design}
\label{sub:transmitter}
In our system, the training phase is organized into \emph{rounds}, each lasting $T$ slots.
At each slot $t$, the transmitter can directly observe the state $s_t$, while the reward $r_t$ and the action $a_t$ are only known to the receiver.
The goal of the transmitter is to build an encoding function $\lambda$ that allows the receiver to efficiently monitor the environment.
To this end, the transmitter takes advantage of the \gls{sfr} scheme presented in Sec.~\ref{sec:homomorphic}, learning two functions, $\phi$ and $\alpha$, to encode the state and the action into a set of features, respectively.

In the rest of the section, we indicate by $\mb{z}_{s}(t)=\phi(s_t) \in \mathbb{R}^F$ and $\mb{z}_{a}(t)=\alpha(a_t) \in \mathbb{R}^A$ the feature vectors for state $s_t$ and action $a_t$, respectively.
Moreover, we indicate by 
\begin{equation}
\label{eq:sfr_state_action}
    \mb{z}_{s,a}(t) = \begin{pmatrix}\phi(s_t)&\alpha(a_t)\end{pmatrix}
\end{equation}
the compound state-action vector.

At each slot $t$, the transmitter observes the state $s_t$ and computes the encoded state representation $\mb{z}_{s}(t)$, which is forwarded to the receiver through the communication channel.
During the same slot, the receiver updates its belief about the current state $s_t$ (as we will detail in Sec.~\ref{sub:receiver}) and takes an action $a_t$, which brings the system to the next state $s_{t+1}$.
Finally, over the entire round, the receiver collects the history of rewards $r_{(n-1)T+1:nT}$ and actions $a_{(n-1)T+1:nT}$, which are inaccessible to the transmitter.

At the end of the $n$-th round, the receiver sends back to the transmitter the rewards $r_{(n-1)T+1:nT}$ and
all the actions $a_{(n-1)T+1:nT}$ taken in that round, in a feedback message.
Hence, after $n$ rounds, the data available to the transmitter is $\mathcal{D}_n=\{d_1,\ldots,d_{nT}\}$, where $d_t=\langle s_t,a_t,r_t,s_{t+1}\rangle$ is the classical experience sample used in \gls{rl}, which contains the state-action pair $(s_t, a_t)$ and the subsequent outcome, i.e., the immediate reward $r_t$ and the next state $s_{t+1}$. 

At the end of the $n$-th round, the transmitter can build the compound representation $\mb{z}_{s,a}(t) $ for each state-action pair in $\mathcal{D}_n$, and uses this information to jointly learn the encoding functions $\phi$ and $\alpha$ as well as the model parameters.
In particular, the transmitter models the environment via the transition probability matrix $\mb{M}\in{\mathbb{R}^{(F+A) \times F}}$, such that
\begin{equation}
    \phi(s_{t+1}) = \mb{z}^\intercal_{s,a}(t)\mb{M},
\end{equation}
and the feature vector $\mb{w}$, such that
\begin{equation}
    r(s_t, a_t) = \mb{z}_{s,a}^\intercal (t) \mb{w}.
\end{equation}

Practically, the learned variables $\phi$, $\alpha$, $\mb{M}$, and $\mb{w}$ are approximated by a \gls{dnn} associated with a parameter vector $\bm{\theta}$.
Therefore, at the end of the $n$-th round, the goal of the transmitter is to tune $\bm{\theta}$ to solve the following optimization problem:
\begin{equation}
    \min_{\phi,\alpha,\mb{w},\mb{M}} \sum_{d_t\in \mc{D}_n}|| \phi(s_{t+1})-\mb{z}^\intercal_{s,a}(t)\mb{M}||_2^2 + (r_t -\mb{w}^\intercal \mb{z}_{s,a}(t))^2.
\end{equation}

A common problem of the above approach is the collapse of the \gls{sfr} space~\cite{balestriero2024learning}. 
To mitigate this risk, \gls{hr3l} implements two distinct state embedding functions $\phi$ and $\phi^-$. 
The first, named \emph{update model}, is used to choose new actions, while the second, named \emph{target model}, is used to predict future Q-values. 
This technique, typically used in \gls{dqn}, has proven to be effective when the target of the loss function is also learned during the optimization procedure.

At the end of each training round, the parameters $\bm{\theta}^-$ of the target model are updated using an exponential moving average with a discount factor $\rho \in (0,1)$:
\begin{equation}
    \bm{\theta}_n^- \gets (1-\rho)\bm{\theta}_n + \rho \bm{\theta}_{n-1}^-\,.
\end{equation}

In addition, we implement the \emph{eligibility trace} method~\cite{schulman2015high}, using a trajectory of subsequent transitions with horizon $H$ for each experience sample in $\mathcal{D}_n$.
This technique reduces compounding errors and makes the overall training phase more robust.
Hence, at the end of the $n$-the round, the loss $\mc{L}_{\bm{\theta}}$ for the transmitter with respect to the parameters $\bm{\theta}$ is 
\begin{equation}
\label{eq:tx_loss}
    \sum_{\mathcal{D}_n}\sum_{h=1}^H ||\phi^-(s_{h+1}) - \mb{y}^\intercal_{s,a}(h)\mb{M}||_2^2 + (r_h - \mb{y}_{s,a}(h)^\intercal\mb{w})^2, 
\end{equation}
where $\mb{y}_{s,a}(h) = \begin{pmatrix} \mb{y}_{s,a}(h-1)^\intercal \mb{M} & \alpha(a_h) \end{pmatrix} $ and
$\mb{y}_{s,a} (1) = \begin{pmatrix}\phi(s_1) &\alpha(a_1) \end{pmatrix} $.


We observe that, assuming that each feature in $\mb{z}_{s}(t) \in \mathbb{R}^F$ is encoded with $L_f$ bits, the uncompressed message carrying the whole state feature vector has a size $|m_t| = F L_f$ bits, which may exceed the channel capacity (defined as the total number of bits that can be transmitted in a time slot).
In this case, \gls{hr3l} resorts to puncturing, selecting only $G_t$ features from $F$ for transmission.
The puncturing operation is represented by a binary vector $\mb{g}_t\in\{0,1\}^F$, where $1$s indicate the selected features, and $0$s the dropped ones.

To fit in the channel capacity limit, the number $G_t$ of selected features must be upper bounded by
\begin{equation}
    G_t=\sum_{f=1}^F g_{t,f}\leq\frac{L}{L_f}-F\;,
\end{equation}
where the $-F$ term accounts for the space needed to include the puncturing mask in the transmitted message, to enable correct decoding at the receiver.
The objective of the transitter is thn to transmit the features that would be more difficult for the receiver to correctly predict.
A possible strategy is hence to pick the $G_t$ features that maximize the function 
\begin{equation}
    \ell(t) = \mb{g}_{t}^\intercal\left(\mb{z}_s(t) - \hat{\mb{z}}_s(t)\right)^2 
\end{equation}
where $\hat{\mb{z}}_s(t) = \mb{z}^\intercal_{s,a}(t-1)\mb{M}_n$ is the feature vector that the receiver could estimate from past information. 
Note that the evaluation of $\hat{\mb{z}}_s(t)$ does not need to be perfect, as an error in the uncertainty of the features will still lead the transmitter to send useful information.

\subsection{Receiver Design} 
\label{sub:receiver}
The receiver's goal is to build a control function $\pi$ that operates over the \gls{sfr} space encoded at the transmitter side.
In this work, we implement \gls{ppo}~\cite{schulman2017proximal} to estimate the optimal control function $\pi^*$, mainly due to its \emph{on-policy} nature.
We recall that on-policy approaches estimate future rewards directly from exploration actions, allowing for a more robust training phase. 
The drawback is that \gls{ppo} and other on-policy algorithms tend to converge to more conservative and therefore less efficient solutions~\cite{chen2023sufficiency}.
However, since our focus is more on robustness than on performance under ideal conditions, on-policy algorithms have an advantage. 

After the $n$-th training round, the transmitter sends the new transition matrix $\mb{M}_n$ and action embedding function $\alpha_n$ to the receiver.
If a state update is lost, the receiver can then update the state-action vector using $\mb{M}_n$, as in \gls{mbrl}: 
\begin{equation}\label{eq:prior_update}
    \hat{\mb{z}}_{s,a}(t)=\begin{pmatrix}
        \hat{\mb{z}}_{s,a}^\intercal(t-1)\mb{M}_n, & \alpha_n(a_{t})
    \end{pmatrix}
\end{equation}
On the other hand, if the message is received successfully, potentially with a delay $d$, the receiver updates its knowledge of the environment using the information contained in the packet. We consider packets to be timestamped, so that the receiver can estimate the delay of each update.

The \emph{posterior} estimate at time $t-d$ of the state features is then obtained as 
\begin{equation}
    \hat{\mb{z}}^*_s(t-d) = \mb{g}_{t-d} \otimes \mb{z}_s(t-d) + (\mb{1}-\mb{g}_{t-d}) \otimes \hat{\mb{z}}_s(t-d),
\end{equation}
where $\otimes$ represents the element-wise product between the vectors. 
Subsequently, the \emph{prior} estimate $\hat{\mb{z}}_s(t)$ at time $t$ is obtained by first evolving the posterior estimate as $\hat{\mb{z}}_s(t+1-d)=\begin{pmatrix} \hat{\mb{z}}_s^*(t-d) & \alpha(a_{t-d}) \end{pmatrix}^\intercal\mb{M}_n$ and then applying the function in~\eqref{eq:prior_update} recursively:
\begin{equation}
   \hat{\mb{z}}_s^*(t-\tau)=\begin{pmatrix} \hat{\mb{z}}_s^*(t-\tau-1) & \alpha(a_{t-\tau-1}) \end{pmatrix}^\intercal\mb{M}_n,
\end{equation}
with $\tau\in\{0,\ldots,d-2\}$. 

The receiver takes advantage of this scheme to maintain a reliable estimate of the current state embedding $\hat{\mb{z}}_s(t)$ and learn a reliable policy $\pi: \mathbb{R}^F \rightarrow \Delta(\mc{A})$.
Following the \gls{ppo} training scheme, the receiver maintains a \gls{rb} $\mc{E}=\{e_1,\ldots,e_T\}$ of experience samples $e_t=\langle(\hat{\mb{z}}_s(t),a_t,r_t,\hat{\mb{z}}_s(t+1))\rangle$ and updates the policy using the clipping technique to avoid catastrophic updates~\cite{schulman2017proximal}. 

We note that other \gls{rl} algorithms, whether on-policy or off-policy, could be employed, rendering our approach agnostic to the specific algorithm used on the receiver side.
On the other hand, \gls{hr3l} requires the receiver to reset its transition buffer at the beginning of each round, as the encoding policy undergoes changes.
Consequently, methods that train the policy based on a single \gls{rb} of experience samples, such as \gls{ppo}, may have better outcomes, as they are more likely to converge faster under the \gls{hr3l} conditions.

\begin{figure*}
    \centering
    \begin{tikzpicture}[
  block/.style = {draw, very thick, rounded corners, minimum width=2cm, minimum height=1cm, align=center},
  node distance=1.1cm
]

\node[block] (tx_enc) {\scriptsize Feature extraction $\phi_n(s_t)$ \\ \scriptsize and compression $\mb{g}_t$};
\node[left=of tx_enc] (input) {};
\node[block, below=0.6cm of tx_enc] (tx_srv) {\scriptsize Learn SFR $\phi_n(\cdot), \alpha_n(\cdot), \mb{M}_n,\mb{w}_n$\\ \scriptsize from $\mathcal{D}_n=\{d_1,\ldots,d_{Tn}\}$};
\node[block, right=1.5cm of tx_enc] (ch) {\scriptsize Wireless channel \\ \scriptsize (delay and loss)};
\node[block, right=1.5cm of ch] (rx) {\scriptsize Latent state estimation \\ \scriptsize using $\alpha_n(\cdot),\mb{M}_n$};
\node[block, right=of rx] (pi) { \scriptsize PPO agent\\ \scriptsize learns $\pi(\bm{\hat{z_t}})$};

\draw[dotted, rounded corners, very thick] (-2.65,0.6) rectangle (2.65,-2.25);
\draw[dotted, rounded corners, very thick] (6.25,0.6) rectangle (12.5,-0.6);

\node[draw,fill=white!80!black,minimum width=2.25cm] (transmitter) at (0,1) {Transmitter};
\node[draw,fill=white!80!black,minimum width=2.25cm] (receiver) at (9.375,1) {Receiver};

\draw[->, line width=1pt] (input) -- node[above, near start] {\scriptsize $s_t$} (tx_enc);
\draw[->, line width=1pt] (tx_enc) -- node[above,near start] {\scriptsize $m_t$} (ch);
\draw[->, line width=1pt] (ch) -- node[above] {\scriptsize $m_{t-d}$} (rx);
\draw[->, line width=1pt] (rx) -- node[above] {\scriptsize $\bm{\hat{z}}_t$} (pi);
\draw[->, dashed, line width=1pt] (tx_srv) -- node[right] {\scriptsize $\phi(\cdot)$} (tx_enc);
\draw[->, dashed, line width=1pt] (pi) |- node[below, near end] {\scriptsize  $a_{(n-1)T:nT-1},r_{(n-1)T:nT-1}$} ([yshift=-0.5cm]tx_srv);
\draw[->, dashed, line width=1pt] ([yshift=0.5cm]tx_srv) -| node[above left] {\scriptsize $\alpha_n(\cdot), \mb{M}_n$} (rx);

\draw[minimum height=1cm,minimum width=2cm,fill=white!80!black] (8.5,-1) rectangle (13.5,-2.1);
\draw[->, line width=1pt] (8.8,-1.3) -- (9.8,-1.3);
\draw[->, dashed, line width=1pt] (8.8,-1.8) -- (9.8,-1.8);
\draw[-] (10,-1.3) -- node[right] {Immediate update} (10,-1.3);
\draw[-] (10,-1.8) -- node[right] {Update Every $T$ steps} (10,-1.8);

\end{tikzpicture}\vspace{-0.8cm}
    \caption{The proposed \gls{hr3l} training architecture.}
    \label{fig:architecture}
\end{figure*}

\begin{figure}[t]
\vspace{-8pt}
\begin{algorithm}[H]
\caption{HR3L Training}
\label{alg:hr3l}
\begin{algorithmic}[1]
\footnotesize

\State Initialize $\mathcal{D} = \{\emptyset\}$
\State Initialize $\mathrm{RB}=\{\emptyset\}$
\For{$n \in 1:N$} \Comment{Repeat for N rounds}
    \For{$t \in 1:T$} 
        \State $\mb{z_s} \leftarrow \phi(s_t)$ 
        \Comment{Transmitter encodes current state}
        \State $\mc{D}\mathrm{.add}(s_t,s_{t+1})$ \Comment{Store Transition in the Dataset}
        \State $m_t \leftarrow \mb{z}_t, \mb{g}$ \Comment{Encode and compress the embedded state}
        \State $\mb{\hat{z}_s} \leftarrow \mathrm{dec}(m_t,h_t)$ \Comment{Decode and estimate the embedded state} 
        \State $a_t \leftarrow \pi(\mb{z_s})$ \Comment{Sample an action}
        \State Collect $r_t$ 
        \State $\mathrm{RB.add}(\mb{\hat{z}_s}(t),a_t,r_t)$   \Comment{Add to Rollout Buffer}
    \EndFor    
    \State Receiver sends $a_{(n-1)T:nT-1},r_{(n-1)T:nT-1}$ to Transmitter 
    \State Receiver updates $\pi$ according to PPO loss
    \State Transmitter updates $\phi_n, \alpha_n, \mb{M}_n, \mb{w}_n$ 
    \State Transmitter sends $\mb{M}_n$ and $\alpha_n$ to Receiver
\EndFor
\end{algorithmic}
\end{algorithm}
\end{figure}

\subsection{Summary of the Training Process}
For the sake of clarity, we conclude this section with a bird's eye view of the framework, considering the main components and steps. The blocks of the \gls{hr3l} architecture are illustrated in Fig.~\ref{fig:architecture}, which provides an overview of the different tasks, from sensor readout at the transmitter to action selection at the receiver, as well as the information flows between the different learning actors. 
Solid and dashed lines denote transmissions that occur at each time slot $t$ and at the end of each training round, respectively.
The sequence of training operations is described by the pseudo-code in Alg.~\ref{alg:hr3l}.

We assume that a full training phase includes a total of $N$ rounds.
Each round lasts for $T$ slots and, at each slot, the transmitter processes the current state $s_t$ through $\phi_n$, obtaining the state embedding $\mb{z}_s(t)$.
The latter constitutes the message $m_t$ that is sent to the receiver, except for the features that are punctured through the binary mask $\mb{g} \in \{0,1\}^F$.
Therefore, the combination of $\mb{g}$ and $\phi_n$ corresponds to the encoding function $\lambda$ given in Sec.~\ref{subsec:rmdp}.
The message $m_t$ is sent to the receiver through the communication channel, which can introduce delay and packet losses, following the model given in Sec.~\ref{subsec:cc}. 
Finally, the receiver uses $\hat{m}_t$ to estimate the current state embedding $\hat{\mb{z}}_s(t)$, and samples a new action according to the policy $\pi$.

At the end of the $n$-th training round, the receiver sends the collected actions and rewards to the transmitter, which performs a new training step, updating the state embedding function $\phi_n$, the action embedding function $\alpha_n$, the transition matrix $\mb{M}_n$, and the reward vector $\mb{w}_n$. 
Hence, the transmitter sends the action embedding function $\alpha_n$ and the transition matrix $\mb{M}_n$ to the receiver, allowing the latter to compensate for message losses and delays. 
It should be noted that the state embedding function $\phi_n$ does not need to be communicated, as the receiver operates over the space given by the \gls{sfr} features and not on the original state space. 

The described procedure is slightly adjusted if the communication channel requires the use of the encoding mask $\mb{g}$ to reduce the number of features transmitted.
In this case, as explained in Sec.~\ref{sub:transmitter}, the transmitter is updated with the new control action $a_t$ at the end of each slot $t$. 
This limitation affects only bandwidth-constrained scenarios and can be readily mitigated in practical deployments, as control actions, unlike states, usually require only a few bytes for encoding.
We finally observe that the \gls{sfr} space evolves simultaneously with the control policy $\pi$, which may introduce instability into the training process.
This problem can be mitigated by properly selecting $T$, allowing the receiver to operate in a stable environment while the encoder refines its feature extraction strategy.

\section{Simulation Settings and Results}
\label{sec:results}

In this section, we describe the setting used to assess the performance of the proposed \gls{hr3l} architecture against two different state-of-the-art approaches.
Our \gls{hr3l} architecture was tested in $25$ different environments from the DeepMind Control Suite~\cite{tassa2018deepmind}, which is a gold standard for evaluating \gls{rl} algorithms, as it comprises a diverse set of environments for learning control policies in robotic tasks.
These tasks span from the classical problem of balancing a cart-pole system to the control of a walking humanoid.

To assess the impact of packet losses and transmission delays, without capacity problems, we considered IoT-like scenarios, where the state $s_t \in \mc{S}$ is a vector of sensor readouts.
In this case, the state-encoding function $\phi$ is approximated by a \gls{fnn}. 
Instead, to evaluate the effects of bandwidth limitation, we focused on image-processing scenarios, assuming that $s_t$ is a sequence of three images.
This is a standard approach when using image observations in \gls{rl}~\cite{mnih2015human} as stacking multiple frames preserves the Markov property of the observations and helps the agent capture temporal context.
In this case, we used a \gls{cnn} with a linear layer at the output as the approximation function $\phi$.
Table~\ref{tab:encoder} reports the details of the neural network architecture in the two cases.

\begin{table}[t]
    \centering
    \caption{Architecture of the agent neural networks}
    \begin{tabular}{c|c|c|c}
        Layers & FNN encoder & CNN encoder & Receiver\\
        \hline
        Input & Lin. $|\mc{S}|\!\times\!128$ & Conv., $32$ ch., RF 3 & Lin. $F\!\times\!256$\\
        Hidden & Lin. $128\!\times\!128$ & 3$\!\times\!$Conv., $32$ ch., RF 3 & Lin. $256\!\times\!256$ \\
        Output & Lin. $128\!\times\!50$ & Lin. $1568\!\times\!512$  & Lin. $256\!\times\! |\mc{A}|$ \\
    \end{tabular}
    \label{tab:encoder}
\end{table}
\begin{table}[b]
    \centering
    \caption{Training parameters of the RL agents}
    \begin{tabular}{c|c|c|c}
        \multicolumn{2}{c|}{\textbf{Receiver (PPO)}} & \multicolumn{2}{c}{\textbf{Transmitter}} \\   
        Parameter & Value & Parameter & Value \\
        \hline
        Learning rate & $3\times10^{-4}$ & Learning rate & $10^{-4}$ \\
        Rollout horizon & 4096 & Horizon & 5 \\
        Batch size & 256 & Batch size & 256 \\
        Epochs & 10 & Epochs & 250 \\
        SDE sampling rate & 4  & Buffer size & $5\times10^5$ \\
    \end{tabular}
    \label{tab:ppo}
\end{table}

To implement the receiver, we adopted the \gls{ppo} implementation from the \verb|StableBaselines3| Python package, whose details are given in~\cite{raffin2021stable}.
In particular, we set the \emph{rollout horizon} to the duration $T$ of a training round, ensuring that the receiver only uses the state embeddings generated with the most recent version of the encoding function.
To implement the transmitter, we used an \emph{ad hoc} Python implementation designed to minimize the loss function in~\eqref{eq:tx_loss}.
The learning parameters used for receiver and transmitter training are reported in Table~\ref{tab:ppo}.  

We then illustrate the results obtained by the considered architecture in the different learning environments provided by the DeepMind Control Suite, under ideal and non-ideal transmission.
Importantly, for each learning problem, we consider three possible communication impairments: transmission delay, packet loss, and bandwidth limitation. 

\subsection{Ideal Communication}

To set a baseline for our analysis, we first consider the case of an ideal communication channel, that is, with no packet losses, delay, or capacity constraints.
We evaluate the benefits of the proposed \gls{hr3l} architecture in terms of \emph{sample efficiency}, i.e., the number of training steps required to achieve a certain performance.
Fig.~\ref{fig:learning_curves} shows the learning curves, averaged across all the learning tasks, for both \gls{hr3l} and a \emph{legacy} \gls{ppo} configuration that does not use \gls{sfr} features, but operates over the original state space.

As shown in the figure, we have considered a total of $2\cdot 10^6$ steps for the training phase. 
We can observe that \gls{hr3l} achieves an average performance higher than that of the legacy \gls{ppo} at each step, showing a higher learning efficiency.
The performance gap is maintained during the testing phase (not shown), where \gls{hr3l} improves the normalized cumulative reward by $12.93\%$ over \gls{ppo}, confirming the advantage of our technique in the case of ideal data transmission. 

\begin{figure}
    \centering
    \begin{tikzpicture}
  \begin{axis}[
      height=4cm,
      width=\linewidth,
      xlabel={Steps [M]},
      ylabel={Normalized Reward},
      legend pos=north west,
      grid=major,
      xmin=0, xmax=1,
      ymin=0, ymax=1,
      xtick={0,0.25,0.5,0.75,1},
      xticklabels={0,0.5,1,1.5,2}
    ]

    \addplot[
      color=color4,
      mark=triangle,
      mark repeat=10,
      very thick
    ] table {figures/learning_curves/r3l.dat};
    \addlegendentry{HR3L}

    \addplot[
      color=color0,
      very thick,
      mark=o,
      mark repeat=10,
      mark options={solid}
    ] table {figures/learning_curves/ppo.dat};
    \addlegendentry{PPO}

  \end{axis}
\end{tikzpicture}
    \vspace{-1cm}
    \caption{Learning curve of HR3L, averaged over the $25$ tasks in the DeepMind Control Suite, under ideal channel conditions.}
    \label{fig:learning_curves}
\end{figure}

\subsection{Impact of Packet Losses}
Here, we evaluate the reliability of \gls{hr3l} in the presence of packet losses, considering a linear state representation and the \gls{fnn} architecture. 
To emulate packet losses, we implement the Gilbert-Elliott model with a transition matrix between good and bad states given by
\begin{equation*}
    P= \begin{pmatrix}
        0.99 & 0.01 \\ 0.1 & 0.9
    \end{pmatrix}.
\end{equation*}
The packet loss probability in the good state is $0.01$.
Instead, for the bad state, we consider two cases, named GE~$95.5$ and GE~$92.5$, with a packet loss probability equal to $0.4$ and $0.7$, resulting in an average success probability of $95.5\%$ and $92.5\%$, respectively.

As an example, in Fig.~\ref{fig:ge} we report a realization of the packet loss pattern for a sequence of $1000$ steps for each model.
We can appreciate that the second channel model is characterized by heavier blockage conditions as the number of losses consistently increases.
Long bursts of losses are particularly harmful for real-time control applications and need to be carefully managed.

\begin{figure}[t]
    \centering
    \begin{tikzpicture}
  \begin{axis}[
      height=3cm,       
      width=\linewidth,
      xlabel={GE 95.5},
      legend pos=north west,
      xmin=0, xmax=1000,
      ymin=0, ymax=1,
      xtick={0,500,1000},
      ytick={0,1},
    ]

    \addplot[
      color=black
    ] table {figures/gilbert_elliott_channels/ge955.dat};

  \end{axis}
\end{tikzpicture}
    \vspace{-1cm}
    \begin{tikzpicture}
  \begin{axis}[       
      height=3cm,       
      width=\linewidth,
      xlabel={GE 92.5},
      legend pos=north west,
      xmin=0, xmax=1000,
      ymin=0, ymax=1,
      xtick={0,500,1000},
      ytick={0,1},
    ]

    \addplot[
      color=black,
    ] table {figures/gilbert_elliott_channels/ge925.dat};

  \end{axis}
\end{tikzpicture}
    \vspace{-1cm}
    \caption{Patterns of the Gilbert-Elliott channel models.}
    \label{fig:ge}
\end{figure}

\begin{figure}
    \centering
    \begin{tikzpicture}
  \begin{axis}[
      ybar,
      height=4cm,
      width=\linewidth,
      symbolic x coords={GE 95.5,GE 92.5},
      xtick=data,
      ylabel={Reward diff. (\%)},
      xlabel={Channel Model},
      legend pos=south west,
      bar width=16pt,
      ymin=-6,
      ymax=0,
      nodes near coords,
      enlarge x limits=0.5,
      grid
    ]
    \addplot+[color4, draw=black, text=black, bar shift=-12.5pt] coordinates {(GE 95.5,-1.2583483773509667) (GE 92.5,-1.6643749311781533)};
    \addplot+[color0, draw=black, text=black, bar shift=12.5pt] coordinates {(GE 95.5,-1.3328121662100019) (GE 92.5,-3.6511373928335353)};
    \legend{HR3L, PPO}
  \end{axis}
\end{tikzpicture}
    \vspace{-1cm}
    \caption{Relative performance loss with respect to the same model under the two packet loss models.}
    \label{fig:packet_loss_perf}
\end{figure}

\begin{figure*}
    \centering
    \begin{tikzpicture}
  \begin{axis}[
      width=\linewidth,
      height=4cm,
      ybar,    
      xlabel style={font=\footnotesize\color{white!15!black}},
      ylabel style={font=\footnotesize\color{white!15!black}},
      tick label style={font=\scriptsize\color{white!15!black}},
      xtick ={0,1,2,3,4,5,6,7,8,9,10,11,12,13,14,15,16,17,18,19,20,21,22,23,24},
      xticklabels={acrobot-swingup,\textbf{ball-in-cup-catch},\textbf{cartpole-balance},\textbf{cartpole-balance-sparse},\textbf{cartpole-swingup},\textbf{cartpole-swingup-sparse},\textbf{cheetah-run},\textbf{dog-run},\textbf{dog-stand},dog-trot,\textbf{dog-walk},{finger-spin},finger-turn-easy,\textbf{finger-turn-hard},\textbf{fish-swim},\textbf{hopper-hop},\textbf{hopper-stand},\textbf{humanoid-run},\textbf{humanoid-stand},\textbf{humanoid-walk},\textbf{pendulum-swingup},\textbf{quadruped-run},\textbf{quadruped-walk},\textbf{reacher-easy},\textbf{reacher-hard}},
      ylabel={Reward reduction (\%)},
      bar width=3pt,
      ymin=0,
      ymax=50,
      xticklabel style = {rotate=30, anchor=east},
      enlarge x limits=0.05,
      grid = major,
      bar width=1pt,
      legend style={at={(0.5,1.25)},font={\scriptsize}, anchor=north,legend columns=6},
    ]

    \draw[c14, thick, on layer=axis foreground] (-2,7.9186299) -- (27,7.9186299);
    \draw[c24, thick, on layer=axis foreground] (-2,8.28684852) -- (27,8.28684852);
    \draw[c34, dashdotted,thick,  on layer=axis foreground] (-2,26.10308621) -- (27,26.10308621);
    \draw[c44, dashdotted, thick, on layer=axis foreground] (-2,27.82717382) -- (27,27.82717382);
    
    \addplot+[c14] table[x=id,y=r2l_955] {figures/comparison_ploss.dat};
    \addplot+[c24] table[x=id,y=r2l_925] {figures/comparison_ploss.dat};
    \addplot+[c34] table[x=id,y=baseline_955] {figures/comparison_ploss.dat};
    \addplot+[c44] table[x=id,y=baseline_925] {figures/comparison_ploss.dat};

    \legend{HR3L (GE 95.5), HR3L (GE 92.5), PPO (GE 95.5), PPO (GE 92.5)};
  \end{axis}
  
\end{tikzpicture}
    \vspace{-1cm}
    \caption{Reward reduction for each control task with respect to the best-performing model as a function of the channel model. The lines indicate the average value across all tasks, and tasks in which \gls{hr3l} outperforms \gls{ppo} are in bold letters.}
    \label{fig:ploss_all_env}
\end{figure*}

Fig.~\ref{fig:packet_loss_perf} shows the degradation in performance of \gls{hr3l} and \gls{ppo} under different packet loss conditions, averaged over all control tasks.
Hence, for each learning architecture and channel conditions, the relative performance loss is computed with respect to the same architecture working in the ideal scenario.
We observe that \gls{hr3l} is robust, limiting the relative performance loss even in the GE~$92.5$ scenario.
On the other hand, \gls{ppo} can handle a small number of packet losses, but its performance severely degrades with GE~$92.5$.
This is because the \gls{ppo} receiver does not create an explicit world model and needs to learn the policy in the environment itself, implicitly reconstructing missing updates.
Instead, the \gls{hr3l} architecture allows the receiver to accurately estimate the \gls{sfr} even after missing an update by using the transition model $\mb{M}_n$.

To allow for a more thorough analysis, we also compute the reward reduction of \gls{hr3l} and \gls{ppo} with respect to the best-performing model (operating with an ideal channel) for both GE~$95.5$ and GE~$92.5$.
Fig.~\ref{fig:ploss_all_env} reports the results for each DeepMind environment: the lines indicate the performance averaged across all tasks, with solid lines for \gls{hr3l} and dashed dotted lines for \gls{ppo}, while the tasks in which \gls{hr3l} outperforms \gls{ppo} are marked in bold.

We observe that \gls{hr3l} fails to outperform \gls{ppo} in only $4$ learning environments.
In $3$ of these cases, \gls{hr3l} does not converge, leading to a large gap with respect to legacy \gls{ppo}; instead, in the \emph{finger spinning} task, its performance is only slightly worse than \gls{ppo}. Additional investigations may be necessary to determine whether this issue can be resolved by increasing the number of training steps or by adjusting the number $T$ of slots per round to improve training stability. We highlight that these results were obtained using \gls{hr3l} as a plug-and-play method, without any task-specific adjustments to its parameters or architecture, and further optimization of the model may yield good results even in the $3$ tasks where it fails to converge.
In all the remaining cases, \gls{hr3l} proves to be more robust than the benchmark method.
In detail, in $10$ of the $25$ cases, \gls{ppo} loses more than $30\%$ of its performance in the presence of packet losses.
In contrast, \gls{hr3l} limits its loss to less than $5\%$ in $9$ of those cases, reaching about $20\%$ loss only in the \emph{quadruped run} task, and experiencing a loss larger than $20\%$ only in the $3$ failure-to-converge scenarios mentioned above.

\subsection{Impact of Communication Delays}
As a second impairment, we consider delayed transmission due to, eg., data acquisition, processing, and compression, or medium access and transmission time.
We model the delay in terms of the number of slots of duration $\delta t$ in the Markov process: hence, a delay of $d_t$ maps into $\left\lceil\frac{d_t}{\delta t}\right\rceil$ slots.
We consider both the case with fixed delay $d$, and a more realistic one with stochastic delay, based on communication traces from a 5G application. 
In both cases, we assume that the packets are timestamped, so that the receiver knows the number of steps since the last update of the environment state. 

In the fixed delay scenario, we assume that a time slot has a duration $\delta t=10$~ms and that $d$ can be equal to $1$, $2$, or $3$ slots.
To allow the legacy \gls{ppo} implicitly estimate the evolution of the system, we augment the receiver's observations by including actions $\{a_{t-d}, \ldots, a_{t-1}\}$ in the input to the control policy, in addition to $s_{t-d}$.
This information, which is always available on the receiver side, enables the \gls{ppo} to emulate the prediction process of \gls{hr3l} that leverages the transition model $\mb{M}_n$ to compute the estimates of the embedded states. 

Instead, the stochastic delay scenario is based on measurements obtained on real hardware under emulated 5G wireless communication~\cite{mostafavi2023data}.
We exploit the 5G Software Defined Radio dataset, which contains recordings of the \gls{rtt} of packets transmitted at intervals of $10$ ms.
This scenario represents a proxy of a realistic delay that a communication link can introduce to a computing-enabled base station integrated with our learning architecture.

First, we isolate the effects of robustness, evening out the difference in sample efficiency by reporting the relative performance loss of each model with respect to the same model under ideal communication conditions in Fig.~\ref{fig:delay_perf}.
As before, each model is compared to its equivalent version operating with a zero-delay channel, and the results are averaged across all the DeepMind tasks. 
The results are comparable when the delay is small, while the performance loss of \gls{ppo} is double that of \gls{hr3l} with the steps $d \geq 2$.
In the stochastic delay scenario, this difference is even more striking, as \gls{ppo} degrades significantly, while \gls{hr3l} maintains a performance comparable to the ideal case.

\begin{figure}[t]
    \centering
    \begin{tikzpicture}
  \begin{axis}[
      ybar,
      height=4cm,
      width=\linewidth,
      symbolic x coords={10 ms, 20 ms, 30 ms, Stochastic},
      xtick=data,
      ylabel={Performance (\%)},
      bar width=13pt,
      ymin=-75,
      ymax=0,
      ytick={0,-10,-20,-30,-40,-50,-60,-70},
      nodes near coords,
      enlarge x limits=0.2,
      grid,
      legend style={anchor=south west,at={(0.015,0.04)}}
    ]
    \addplot+[color4, draw=black, text=black, font=\footnotesize, bar shift=-12pt] coordinates {
    (10 ms,-8.186357340313016) 
    (20 ms,-11.243216906471927)
    (30 ms,-15.542341847050384)
    (Stochastic, -19.54)};
    \addplot+[color0, draw=black, text=black, font=\footnotesize, bar shift=12pt] coordinates {
    (10 ms,-10.892821724022873) 
    (20 ms,-21.66633252980997)
    (30 ms,-30.827370452009053)
    (Stochastic, -61.89)
    };
    \legend{HR3L,PPO}
  \end{axis}
\end{tikzpicture}
    \vspace{-1cm}
    \caption{Relative performance loss with respect to the same model considering different communication delays.}
    \label{fig:delay_perf}
\end{figure}

\begin{figure*}
    \centering
    \subfloat[\gls{hr3l}.\label{fig:delay_all_env_hr3l}]
    {\begin{tikzpicture}
  \begin{axis}[
      width=\linewidth,
      height=4cm,
      ybar,    
      xlabel style={font=\footnotesize\color{white!15!black}},
      ylabel style={font=\footnotesize\color{white!15!black}},
      tick label style={font=\scriptsize\color{white!15!black}},
      xtick ={0,1,2,3,4,5,6,7,8,9,10,11,12,13,14,15,16,17,18,19,20,21,22,23,24},
      xticklabels={acrobot-swingup,\textbf{ball-in-cup-catch},\textbf{cartpole-balance},\textbf{cartpole-balance-sparse},\textbf{cartpole-swingup},\textbf{cartpole-swingup-sparse},\textbf{cheetah-run},\textbf{dog-run},\textbf{dog-stand},dog-trot,\textbf{dog-walk},{finger-spin},finger-turn-easy,\textbf{finger-turn-hard},\textbf{fish-swim},\textbf{hopper-hop},\textbf{hopper-stand},\textbf{humanoid-run},\textbf{humanoid-stand},\textbf{humanoid-walk},\textbf{pendulum-swingup},\textbf{quadruped-run},\textbf{quadruped-walk},\textbf{reacher-easy},\textbf{reacher-hard}},
      ylabel={Reward reduction (\%)},
      bar width=3pt,
      ymin=0,
      ymax=100,
      xticklabel style = {rotate=30, anchor=east},
      enlarge x limits=0.05,
      grid = major,
      bar width=1pt,
      legend style={at={(0.5,1.25)},font={\scriptsize}, anchor=north,legend columns=6},
    ]

    \draw[c14, thick, on layer=axis foreground] (-2,13.74) -- (27,13.74);
    \draw[c24, thick, on layer=axis foreground] (-2,16.72) -- (27,16.72);
    \draw[c34, thick, on layer=axis foreground] (-2,20.75) -- (27,20.75);
    \draw[c44, thick,  on layer=axis foreground] (-2,26.88) -- (27,26.88);
    
    \addplot+[c14] table[x=id,y=r2l_1step] {figures/comparison_delay.dat};
    \addplot+[c24] table[x=id,y=r2l_2step] {figures/comparison_delay.dat};
    \addplot+[c34] table[x=id,y=r2l_3step] {figures/comparison.dat};
    \addplot+[c44] table[x=id,y=r2l_stochastic_delay] {figures/comparison_delay.dat};

    \legend{HR3L ($d=10$ms), HR3L ($d=20$ms), HR3L ($d=30$ms), H3RL (stochastic)};
  \end{axis}
  
\end{tikzpicture}}\\
    \subfloat[\gls{ppo}.\label{fig:delay_all_env_ppo}]
    {\begin{tikzpicture}
  \begin{axis}[
      width=\linewidth,
      height=4cm,
      ybar,    
      xlabel style={font=\footnotesize\color{white!15!black}},
      ylabel style={font=\footnotesize\color{white!15!black}},
      tick label style={font=\scriptsize\color{white!15!black}},
      xtick ={0,1,2,3,4,5,6,7,8,9,10,11,12,13,14,15,16,17,18,19,20,21,22,23,24},
      xticklabels={acrobot-swingup,\textbf{ball-in-cup-catch},\textbf{cartpole-balance},\textbf{cartpole-balance-sparse},\textbf{cartpole-swingup},\textbf{cartpole-swingup-sparse},\textbf{cheetah-run},\textbf{dog-run},\textbf{dog-stand},dog-trot,\textbf{dog-walk},{finger-spin},finger-turn-easy,\textbf{finger-turn-hard},\textbf{fish-swim},\textbf{hopper-hop},\textbf{hopper-stand},\textbf{humanoid-run},\textbf{humanoid-stand},\textbf{humanoid-walk},\textbf{pendulum-swingup},\textbf{quadruped-run},\textbf{quadruped-walk},\textbf{reacher-easy},\textbf{reacher-hard}},
      ylabel={Reward reduction (\%)},
      bar width=3pt,
      ymin=0,
      ymax=100,
      xticklabel style = {rotate=30, anchor=east},
      enlarge x limits=0.05,
      grid = major,
      bar width=1pt,
      legend style={at={(0.5,1.25)},font={\scriptsize}, anchor=north,legend columns=6},
    ]

    \draw[c14, dashdotted, thick, on layer=axis foreground] (-2,33.36) -- (27,33.36);
    \draw[c24, dashdotted, thick, on layer=axis foreground] (-2,40.35) -- (27,40.35);
    \draw[c34, dashdotted, thick, on layer=axis foreground] (-2,47.49) -- (27,47.49);
    \draw[c44, dashdotted, thick,  on layer=axis foreground] (-2,79.18) -- (27,79.18);
    
    \addplot+[c14] table[x=id,y=baseline_1step] {figures/comparison_delay.dat};
    \addplot+[c24] table[x=id,y=baseline_2step] {figures/comparison_delay.dat};
    \addplot+[c34] table[x=id,y=baseline_3step] {figures/comparison.dat};
    \addplot+[c44] table[x=id,y=baseline_stochastic_delay] {figures/comparison_delay.dat};

    \legend{PPO ($d=10$ms), PPO ($d=20$ms), PPO ($d=30$ms), PPO (stochastic)};
  \end{axis}
  
\end{tikzpicture}}
    \caption{Reward reduction for each control task with respect to the best-performing model as a function of the communication delay. The lines indicate the average value across all tasks, and tasks in which \gls{hr3l} outperforms \gls{ppo} are in bold letters.}
    \label{fig:delay_all_env}
\end{figure*}

We then analyze the overall performance of the two approaches. Fig.~\ref{fig:delay_all_env_hr3l} shows the degradation of the reward of \gls{hr3l} under the different delay settings with respect to the best-performing model, between \gls{hr3l} and \gls{ppo}, in each of the DeepMind environments.
Our training architecture overcomes \gls{ppo} in $21$ out of $25$ tasks, showing low performance only in the $4$ tasks previously observed as critical.
We also note that the stochastic delay is not significantly more harmful than a fixed delay, except for a few cases (the \emph{cart-pole swing-up}, \emph{hopper hop}, and \emph{pendulum swing-up} tasks).
This further highlights the robustness of \gls{hr3l} to multiple consecutive time slots without updates from the transmitter.

In contrast, legacy \gls{ppo} performs significantly worse than \gls{hr3l}, with an average loss of reward between $35\%$ and $45\%$ for static delay and an average loss of $80\%$ for stochastic delays.
We note that both \gls{ppo} and \gls{hr3l} have been trained in the same environment, so this gap is due to the difficulty of \gls{ppo} in dealing with missing updates, not training issues.

\subsection{Impact of Capacity Limitations}
In this section, we analyze the performance of \gls{hr3l} in the presence of capacity limits of the communication channel, applying the encoding scheme proposed in Sec.~\ref{sub:transmitter}.
We focus on the scenario where the state is derived from visual representations (i.e., images) of the environment, using the \gls{cnn} model at the transmitter side.
In particular, we set the number of features to $F=512$ (which corresponds to the size of the read-out layer of the \gls{cnn} model) and represent each feature with $L_f=16$ bits. 

We note that the DeepMind environments consider $84\times84$ RGB images with 8 bits per channel, for a total data rate of $\approx 17$~Mb/s for raw data transmission. 
Using our \gls{sfr} representation, even without any feature selection ($G=F$), we obtain a data rate $S = |m_t|/ \delta t \approx 0.82$~Mb/s. 
Therefore, by encoding the state into $512$ features, \gls{hr3l} leads to a strong improvement with respect to the transmission of raw data, both in terms of bandwidth occupation and storage requirements.

\begin{figure}
    \centering
    \begin{tikzpicture}
  \begin{axis}[
      height=4cm,
      width=\linewidth,
      xlabel={Steps ($\times10^6$)},
      ylabel={Episode Reward},
      grid=major,
      xmin=0, xmax=1,
      ymin=0, ymax=0.6,
      xtick={0,0.25,0.5,0.75,1},
      xticklabels={0,0.5,1,1.5,2},
      ytick={0,0.2,0.4,0.6},
      yticklabels={0,200,400,600},
      legend style={anchor=east,at={(0.985,0.4)},legend columns=3}
    ]

    \addplot[
      color=color4,
      very thick,
      mark=triangle
    ] table {
    0.0 0.052348
    0.1 0.254589
    0.2 0.353890
    0.3 0.394796
    0.4 0.439854
    0.5 0.479248
    0.6 0.502905
    0.7 0.513098
    0.8 0.514224
    0.9 0.515289
    1.0 0.516348
    };
    \addlegendentry{HR3L}

    \addplot[
      color=color2,
      very thick,
      mark=square
    ] table {
    0.0 0.052938
    0.1 0.284090
    0.2 0.383955
    0.3 0.432818
    0.4 0.482052
    0.5 0.513409
    0.6 0.523249
    0.7 0.514209
    0.8 0.523405
    0.9 0.513892
    1.0 0.529484
    };
    \addlegendentry{DrQv2}

    \addplot[
      color=color0,
      very thick,
      mark=o
    ] table {
    0.0 0.050932
    0.1 0.097648
    0.2 0.121903
    0.3 0.128326
    0.4 0.123059
    0.5 0.119420
    0.6 0.120395
    0.7 0.128326
    0.8 0.125483
    0.9 0.119837
    1.0 0.124895
    };
    \addlegendentry{PPO}

  \end{axis}
\end{tikzpicture}
    \vspace{-1cm}
    \caption{Learning curves for visual \textit{cheetah-run} environment.}
    \label{fig:visual-curves}\vspace{-0.5cm}
\end{figure}

In order to provide a fair comparison, we also implement two state-of-the-art lossy compression techniques as baselines. 
Specifically, we consider the JPEG compression provided by \verb|OpenCV|, one of the most common image processing libraries~\cite{opencv_library}.
This JPEG encoder makes it possible to trade off image data size and distortion through a control parameter in the range $[0,\, 100]$, where $100$ corresponds to the highest image quality level and, consequently, the lowest compression.

As a second, more refined benchmark, we consider the \verb|CompressAI| platform~\cite{begaint2020compressai}, which contains several compression algorithms for images.
Hence, we implement the \gls{dnn} compression technique designed in~\cite{balle2018variational}, which provides eight different trade-offs in terms of data size and distortion. 
In the following, we refer to the latter method as \textit{CAI}. 

We integrate both JPEG and CAI with DrQv2, a baseline \gls{rl} method for 2D observations. 
This method was first proposed in \cite{yarats2021mastering} and has shown high performance in most control tasks in the DeepMind Control Suite, providing better sampling efficiency than traditional \gls{rl} algorithms such as \gls{ppo}. 
In our analysis, DrQv2 is used to learn the control policy $\pi$ while JPEG or CAI is used to compress the environment observations. 
On the other hand, the \gls{hr3l} architecture still adopts \gls{ppo} at the receiver, as the images are encoded in the feature vector on the transmitter side.

\begin{figure}
    \centering
    \begin{tikzpicture}
  \begin{axis}[
      height=4cm,
      width=\linewidth,
      xlabel={Required bitrate [Mb/s]},
      ylabel={Normalized Reward},
      grid=major,
      xmin=0, xmax=0.6,
      ymin=0.5, ymax=1,
      legend style={anchor=south,at={(0.5,0.04)}}
    ]
    \addplot[
      color=color4,
      very thick,
      mark=triangle
    ] table {0.36828875 1.0
0.2461335 0.9838803026775147
0.0824605 0.9502878721227478
0.04174175 0.9034043266037777
0.020983250000000002 0.8444504780827572
0.01699125 0.8229507830050901
0.012201 0.7848999972776877
0.01060425 0.7696832337575187
0.009007250000000001 0.7332905360958418
};
    \addlegendentry{HR3L}
    
    \addplot[
      color=color2,
      very thick,
      mark=square
    ] table {1.1604 1.0
0.8696 0.9374476842337518
0.737494 0.9251099620885028
0.6609205 0.9017687654147812
0.614634 0.8840068176784475
0.5520535 0.7281750316213336
0.48974350000000005 0.5378082947500762
};
    \addlegendentry{DrQv2 (JPEG)}
    
\addplot[
      color=color0,
      very thick,
      mark=o
    ] table {0.0776115 0.7494926263802821
0.1085325 0.8784333223837371
0.1456 0.9828607939233577
0.19705 1.0
};
    \addlegendentry{DrQv2 (CAI)}

  \end{axis}
\end{tikzpicture}
    \vspace{-1cm}
    \caption{Pareto curves of different models between the relative performance and the required bitrate.}
    \label{fig:visual-perf}
\end{figure}

In Fig.~\ref{fig:visual-curves}, we compare the learning curves of \gls{hr3l}, DrQv2, and legacy \gls{ppo} using full-quality image representations.
From the results, we observe that the sample efficiency of \gls{ppo} is rather low: at the end of the training phase, it barely reaches $21\%$ of the performance of \gls{hr3l}, even if no compression has been adopted.
We observe that DrQv2 and \gls{hr3l} lead to similar sample efficiency and, consequently, to approximately the same average reward after $2\times10^6$ training steps.
As both algorithms converge to the same performance, we used it as a normalization factor in the following.

In Fig.~\ref{fig:visual-perf}, we then compare \gls{hr3l} against DrQv2 under capacity limitations, showing the Pareto curves between compression and normalized reward.
Notably, the DrQv2 configuration exploiting CAI has been trained using the same compression settings as in the testing phase, to provide the as fair as possible comparison. 
We observe that JPEG requires a transmission capacity that is, in general, much higher than for the other methods.
On the other hand, \gls{hr3l} and CAI obtain much better results, reaching good performance in terms of reward with much lower data rates.

It should be noted that while DrQv2 with CAI achieves a marginally higher reward at data rates exceeding $\approx0.17$~Mb/s, \gls{hr3l} delivers comparable performance across a significantly broader range of channel capacities.
Furthermore, the proposed method exhibits a graceful performance degradation even as the data rate falls below $\approx 0.07$~Mbit/s. Additionally, contrary to the other techniques, \gls{hr3l} does not need to decompress the images at the receiver, making it suitable for deployment on low-cost devices.
To study this aspect, we measure the delay introduced by compression and decompression (this only in the case of JPEG and CAI) under equal processing capabilities.
In particular, we consider a machine equipped with an Intel Core i7-9700K CPU and a NVIDIA GeForce RTX 2080 Ti GPU.

\begin{figure}
\centering
    \begin{tikzpicture}
  \begin{axis}[
        scale only axis,
        clip=false,
        separate axis lines,
        axis on top,
        height=1.5cm,
        width=7cm,
        xmin=0,
        xmax=4,
        xtick={1,2,3},
        x tick style={draw=none},
        xticklabels={HR3L,CAI,JPEG},
        ymin=0,
        ymax=10,
        ylabel={Delay [ms]},
        xlabel={Compression method},
        grid,
        nodes near coords,
        every axis plot/.append style={
          ybar,
          bar width=0.4,
          bar shift=0pt,
          fill
        }
      ]
      \addplot[color4, draw=black, text=black]coordinates{(1,0.6)};
      \addplot[color2, draw=black, text=black]coordinates{(2,6.1)};
      \addplot[color0, draw=black, text=black]coordinates {(3,1.25)};

  \end{axis}
\end{tikzpicture}
    \vspace{-1cm}
    \caption{Average delay introduced by the compression methods.}
    \label{fig:delay_comp}
\end{figure}

Fig.~\ref{fig:delay_comp} reports the average end-to-end delay incurred by the three methodologies when processing a state image.
We note that \gls{hr3l} introduces the smallest delay, even below that of JPEG, since the receiver operates directly over the \gls{sfr} space.
In contrast, CAI needs to decompress the received images, and its average delay approaches the duration of the timestep $\delta t=10$~ms, an order of magnitude larger than \gls{hr3l}.
This represents a critical weakness of state-of-the-art compression methods: the additional delay due to the compression and decompression pipeline can be comparable to the observation-action loop period, determining a further reduction in performance. 

\begin{figure*}
    \centering
    \pgfplotsset{scaled y ticks=false}
\begin{tikzpicture}
  \begin{axis}[
      width=\linewidth,
      height=4cm,
      ybar,    
      xlabel style={font=\footnotesize\color{white!15!black}},
      ylabel style={font=\footnotesize\color{white!15!black}},
      tick label style={font=\scriptsize\color{white!15!black}},
      xtick ={0,1,2,3,4,5,6,7,8,9,10,11,12,13,14,15,16,17,18,19,20,21,22,23,24},
      xticklabels={acrobot-swingup,\textbf{ball-in-cup-catch},\textbf{cartpole-balance},\textbf{cartpole-balance-sparse},\textbf{cartpole-swingup},\textbf{cartpole-swingup-sparse},\textbf{cheetah-run},\textbf{dog-run},\textbf{dog-stand},\textbf{dog-trot},dog-walk,finger-spin,\textbf{finger-turn-easy},\textbf{finger-turn-hard},fish-swim,hopper-hop,\textbf{hopper-stand},\textbf{humanoid-run},\textbf{humanoid-stand},\textbf{humanoid-walk},\textbf{pendulum-swingup},\textbf{quadruped-run},\textbf{quadruped-walk},\textbf{reacher-easy},\textbf{reacher-hard}},
      ylabel={Average data rate [Mb/s]},
      bar width=3pt,
      ymin=0,
      ymax=0.2,
      ytick={0,0.05,0.1,0.15,0.2},
      yticklabels={0,0.05,0.1,0.15,0.2},
      xticklabel style = {rotate=30, anchor=east},
      enlarge x limits=0.05,
      grid = major,
      bar width=1pt,
      legend style={at={(0.5,1.25)},font={\scriptsize}, anchor=north,legend columns=6},
    ]

    \draw[color4, thick, on layer=axis foreground] (-2,0.1113221360355166) -- (27,0.1113221360355166);
    
    \draw[color2, thick, dashdotted, on layer=axis foreground] (-2,0.1322557) -- (27,0.1322557);
    
    \addplot+[color4] table[x=id,y=R2L] {figures/average_data_rate_per_env.dat};
    \addplot+[color2] table[x=id,y=CompressAI] {figures/average_data_rate_per_env.dat};
    \addplot+[color0] table[x=id,y=JPEG] {figures/average_data_rate_per_env.dat};
    \legend{HR3L, DrQv2 (CAI), DrQv2 (JPEG)};

  \end{axis}
  
\end{tikzpicture}
    \vspace{-1cm}
    \caption{Average bitrate in each environment for the different compression methods while considering a required normalized reward of 0.9. The horizontal lines indicate the average rate across all tasks. The tasks where \gls{hr3l} outperforms DrQv2 are in bold letters.}
    \label{fig:rate_per_env}
\end{figure*}

We then investigate the data rate required to reach $90\%$ of the average reward obtained when using all the $512$ features to represent the state of each control task, or when using DrQv2 with a full-quality image.
The results, shown in Fig.~\ref{fig:rate_per_env}, confirm that JPEG performs worse than the other methods, requiring more than $0.6$~Mb/s to reach the target performance level in all tasks.
In addition, CAI requires a bitrate higher than that of \gls{hr3l} in $20$ of the $25$ tasks, with an average communication overhead (with respect to \gls{hr3l}) of $18.8\%$.

\subsection{Combined Scenario}

As a final experiment, we consider two communication scenarios presenting multiple channel impairments.
The first scenario, named \emph{stationary}, includes the GE~$95.5$ channel model for packet losses, a fixed delay of $d=20$~ms, and a transmission capacity limited to $0.15$~Mb/s.
The second scenario, named \emph{high mobility}, considers stochastic delays and limited capacity, but without an explicit packet loss model.
However, if a packet arrives later than the subsequent update, it is dropped by the receiver.

\begin{figure*}
    \centering
    \begin{tikzpicture}
  \begin{axis}[
      width=\linewidth,
      height=4cm,
      ybar,    
      xlabel style={font=\footnotesize\color{white!15!black}},
      ylabel style={font=\footnotesize\color{white!15!black}},
      tick label style={font=\scriptsize\color{white!15!black}},
      xtick ={0,1,2,3,4,5,6,7,8,9,10,11,12,13,14,15,16,17,18,19,20,21,22,23,24},
      xticklabels={acrobot-swingup,\textbf{ball-in-cup-catch},cartpole-balance,\textbf{cartpole-balance-sparse},{cartpole-swingup},\textbf{cartpole-swingup-sparse},\textbf{cheetah-run},dog-run,\textbf{dog-stand},\textbf{dog-trot},\textbf{dog-walk},{finger-spin},\textbf{finger-turn-easy},\textbf{finger-turn-hard},\textbf{fish-swim},\textbf{hopper-hop},\textbf{hopper-stand},\textbf{humanoid-run},\textbf{humanoid-stand},\textbf{humanoid-walk},\textbf{pendulum-swingup},\textbf{quadruped-run},\textbf{quadruped-walk},\textbf{reacher-easy},{reacher-hard}},
      ylabel={Reward reduction (\%)},
      bar width=3pt,
      ymin=0,
      ymax=80,
      xticklabel style = {rotate=30, anchor=east},
      enlarge x limits=0.05,
      grid = major,
      bar width=1pt,
      legend style={at={(0.5,1.25)},font={\scriptsize}, anchor=north,legend columns=6},
    ]

    \draw[c14, thick, on layer=axis foreground] (-2,19.284091959324712) -- (27,19.284091959324712);
    \draw[c24, thick, on layer=axis foreground] (-2,34.9036) -- (27,34.9036);
    \draw[c34, thick, dashdotted, on layer=axis foreground] (-2,44.37) -- (27,44.37);
    \draw[c44, thick, dashdotted, on layer=axis foreground] (-2,59.39) -- (27,59.39);
    
    \addplot+[c14] table[x=id,y=r2l_mixed] {figures/visual_res.dat};
    \addplot+[c24] table[x=id,y=r2l_stochastic] {figures/visual_res.dat};
    \addplot+[c34] table[x=id,y=baseline_mixed] {figures/visual_res.dat};
    \addplot+[c44] table[x=id,y=baseline_stochastic] {figures/visual_res.dat};

    \legend{HR3L (stationary), HR3L (high mobility), DrQv2 (stationary), DrQv2 (high mobility), };
  \end{axis}
  
\end{tikzpicture}
    \vspace{-1cm}
    \caption{Reward reduction for each control task with respect to the best-performing model as a function of the communication scenario. The lines indicate the average value across all tasks, and tasks in which \gls{hr3l} outperforms DrQv2 are in bold letters.}
    \label{fig:perf_visual_env}
\end{figure*}

We compare \gls{hr3l} with DrQv2 using the CAI scheme, for which we neglected the processing additional delay incurred for image compression and decompression to focus only on the effect of channel impairments.
As DrQv2 does not have a proper way to deal with observation losses, we consider that the benchmark follows an ideal training process that takes place at a centralized node with perfect knowledge of the sequence of state observations.
This assumption may not be verified in practice, making DrQv2 or similar approaches unsuitable in real world scenarios.
In contrast, \gls{hr3l} employs a distributed training strategy by design, allowing it to be trained under the same communication conditions that match those encountered during the final deployment.
 
Fig.~\ref{fig:perf_visual_env} shows the reward loss with respect to the best-performing model for each control task in the two communication scenarios.
Working conditions are much more challenging than in the previous analyzes, as the observations are compressed, delayed, and sometimes erased by the channel, significantly degrading control performance.
On the other hand, in the stationary scenario, \gls{hr3l} limits its loss to $19.3\%$ with respect to the ideal communication case, while DrQv2 performance drops by $44.4\%$, performing worse than \gls{hr3l} in $21$ of $25$ tasks.
The high mobility scenario proves even more difficult.: the average performance of \gls{hr3l} is degraded by $34.9\%$, but still outperforms DrQv2, whose average performance decreases by $59.4\%$, in $19$ tasks.

These results confirm that \gls{hr3l} can allow much higher robustness to channel impairments, while reducing the communication overhead necessary to train \gls{rl} agents in networked systems.
Notably, state-of-the-art methods, although trained \emph{ad hoc} for the target communication channel, fail to make accurate decisions when fresh information is unavailable. 

\section{Conclusions}
\label{sec:conclusion}

In this work, we proposed a novel architecture for training \gls{rrl} agents over unreliable communication channels, addressing several key limitations of such systems.
Our framework, named \gls{hr3l}, takes advantage of homomorphic \gls{mdp} theory to construct goal-oriented representations of environment observations, strongly reducing the communication cost in the case of high-dimensional state spaces.
In addition, \gls{hr3l} is designed to be robust to channel impairments, which can cause state observations to be lost or delayed. To our knowledge, the proposed framework is the first to allow for fully distributed training, avoiding the need for differentiable communication channels or synchronized gradient updates, as required by most state-of-the-art \gls{rrl} methods. 

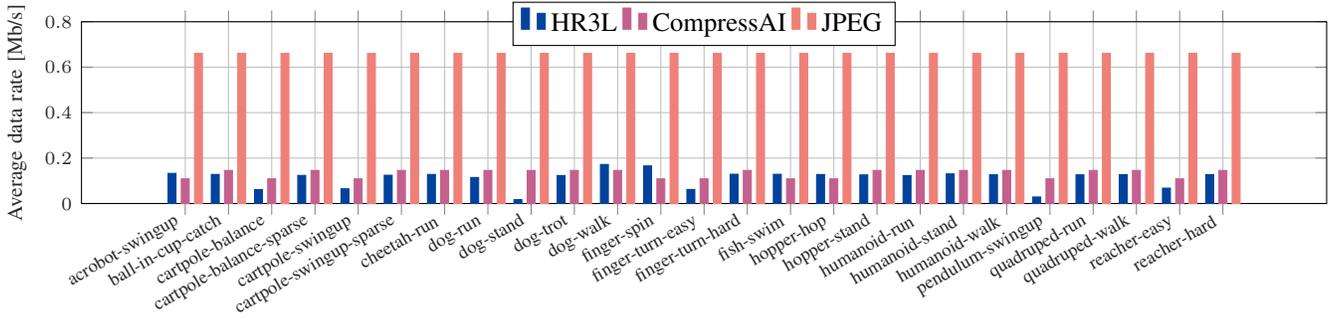
We evaluated the performance of \gls{hr3l} in several \gls{rl} environments of the DeepMind Control Suite and under different communication conditions.
Our analysis against state-of-the-art methods shows that \gls{hr3l} can reduce the required bandwidth for complex observations, mitigate the computational burden on the receiver side, and is significantly more robust to packet loss and transmission delay even with respect to \emph{ad hoc} models trained for specific channels.

There are several potential avenues for future research, such as designing new methods to distribute information between the transmitter and the receiver, particularly when the latter can make its own imperfect state observations.
The extension of these methods to multi-agent cases, which may make our scheme more broadly applicable, is another challenging problem.
Finally, considering the energy limitations and computational capabilities of the transmitter and receiver will be a crucial aspect in realistic resource-constrained settings.

\bibliographystyle{IEEEtran}
\bibliography{biblio}

\end{document}